\documentclass[letterpaper, 10 pt, conference]{ieeeconf}
\IEEEoverridecommandlockouts                              
\usepackage{times}
\usepackage{epsfig}
\usepackage{graphicx}
\usepackage{amsmath}
\usepackage{amssymb}

\usepackage{graphicx}
\usepackage{epsfig}
\usepackage{epic,eepic}
\usepackage{times}
\usepackage{mathptmx}
\usepackage{amsfonts}
\usepackage{amsmath}
\usepackage{cite}
\usepackage{color}

\usepackage{times}
\usepackage{multicol}

\definecolor{red}{rgb}{1,0,0}
\definecolor{green}{rgb}{0,1,0}
\definecolor{blue}{rgb}{0,0,1}
\definecolor{violet}{rgb}{1,0,1}
\definecolor{cyan}{cmyk}{1,0,0,0}
\definecolor{magenta}{cmyk}{0,1,0,0}
\definecolor{yellow}{cmyk}{0,0,1,0}

\definecolor{white}{rgb}{1,1,1}

\newcommand{\CO}[1]{}

\newcommand{\CommentOut}[1]{}

 \newcommand{\editage}[1]{}

\begin{document}

\newcommand{\FIG}[3]{
\begin{minipage}[b]{#1cm}
\begin{center}
\includegraphics[width=#1cm]{#2}\\
{\scriptsize #3}
\end{center}
\end{minipage}
}

\newcommand{\FIGU}[3]{
\begin{minipage}[b]{#1cm}
\begin{center}
\includegraphics[width=#1cm,angle=180]{#2}\\
{\scriptsize #3}
\end{center}
\end{minipage}
}

\newcommand{\FIGm}[3]{
\begin{minipage}[b]{#1cm}
\begin{center}
\includegraphics[width=#1cm]{#2}\\
{\scriptsize #3}
\end{center}
\end{minipage}
}

\newcommand{\FIGR}[3]{
\begin{minipage}[b]{#1cm}
\begin{center}
\includegraphics[angle=-90,width=#1cm]{#2}
\\
{\scriptsize #3}
\vspace*{1mm}
\end{center}
\end{minipage}
}

\newcommand{\FIGRpng}[5]{
\begin{minipage}[b]{#1cm}
\begin{center}
\includegraphics[bb=0 0 #4 #5, angle=-90,clip,width=#1cm]{#2}\vspace*{1mm}
\\
{\scriptsize #3}
\vspace*{1mm}
\end{center}
\end{minipage}
}

\newcommand{\FIGpng}[5]{
\begin{minipage}[b]{#1cm}
\begin{center}
\includegraphics[bb=0 0 #4 #5, clip, width=#1cm]{#2}\vspace*{-1mm}\\
{\scriptsize #3}
\vspace*{1mm}
\end{center}
\end{minipage}
}

\newcommand{\FIGtpng}[5]{
\begin{minipage}[t]{#1cm}
\begin{center}
\includegraphics[bb=0 0 #4 #5, clip,width=#1cm]{#2}\vspace*{1mm}
\\
{\scriptsize #3}
\vspace*{1mm}
\end{center}
\end{minipage}
}

\newcommand{\FIGRt}[3]{
\begin{minipage}[t]{#1cm}
\begin{center}
\includegraphics[angle=-90,clip,width=#1cm]{#2}\vspace*{1mm}
\\
{\scriptsize #3}
\vspace*{1mm}
\end{center}
\end{minipage}
}

\newcommand{\FIGRm}[3]{
\begin{minipage}[b]{#1cm}
\begin{center}
\includegraphics[angle=-90,clip,width=#1cm]{#2}\vspace*{0mm}
\\
{\scriptsize #3}
\vspace*{1mm}
\end{center}
\end{minipage}
}

\newcommand{\FIGC}[5]{
\begin{minipage}[b]{#1cm}
\begin{center}
\includegraphics[width=#2cm,height=#3cm]{#4}~$\Longrightarrow$\vspace*{0mm}
\\
{\scriptsize #5}
\vspace*{8mm}
\end{center}
\end{minipage}
}

\newcommand{\FIGf}[3]{
\begin{minipage}[b]{#1cm}
\begin{center}
\fbox{\includegraphics[width=#1cm]{#2}}\vspace*{0.5mm}\\
{\scriptsize #3}
\end{center}
\end{minipage}
}

\newcommand{\figA}{
\begin{figure}
\FIG{8}{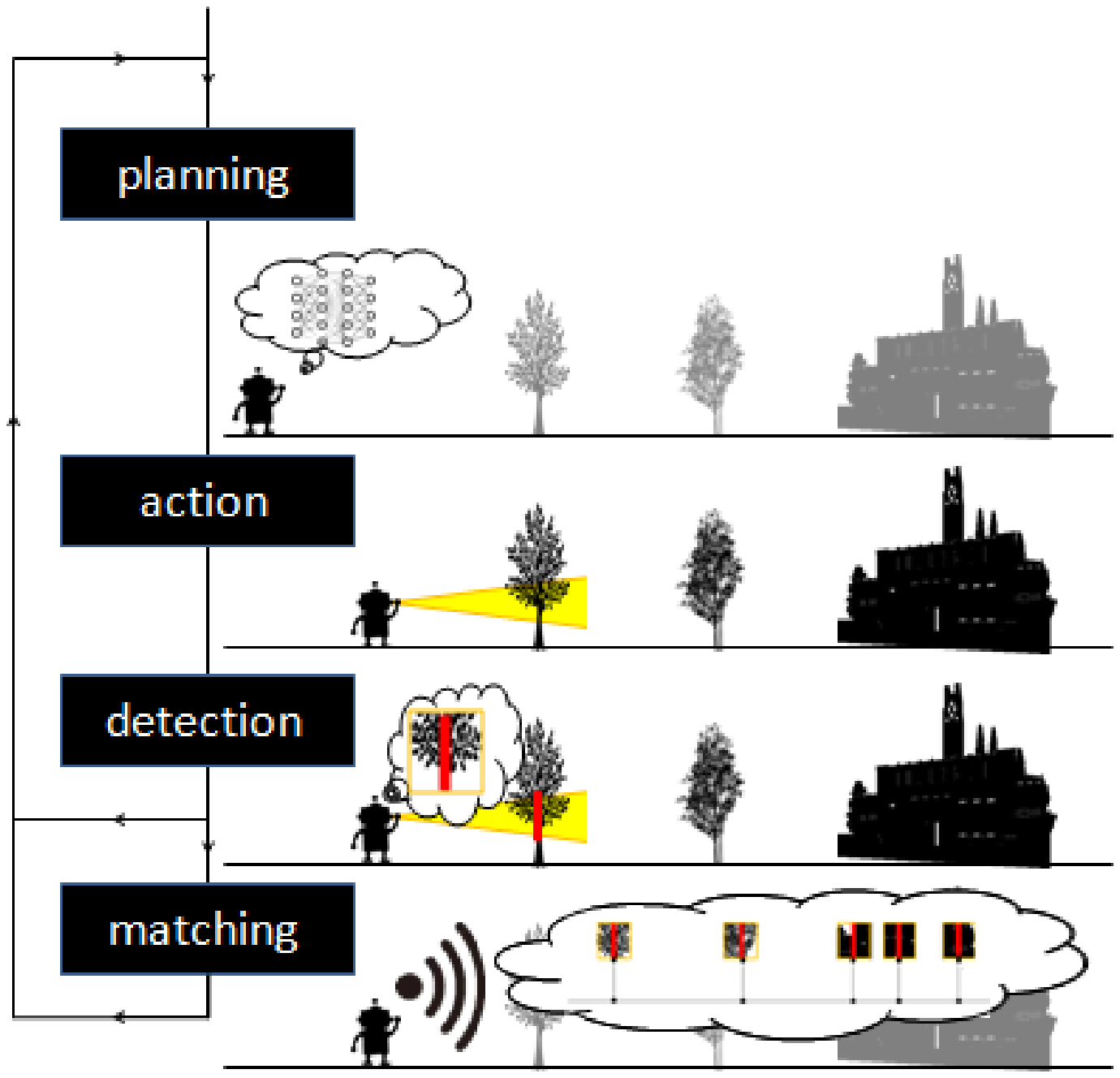}{}
\caption{Active self-localization task.}\label{fig:A}
\end{figure}
}

\newcommand{\figO}{
\begin{figure}
\centering
\FIG{3.3}{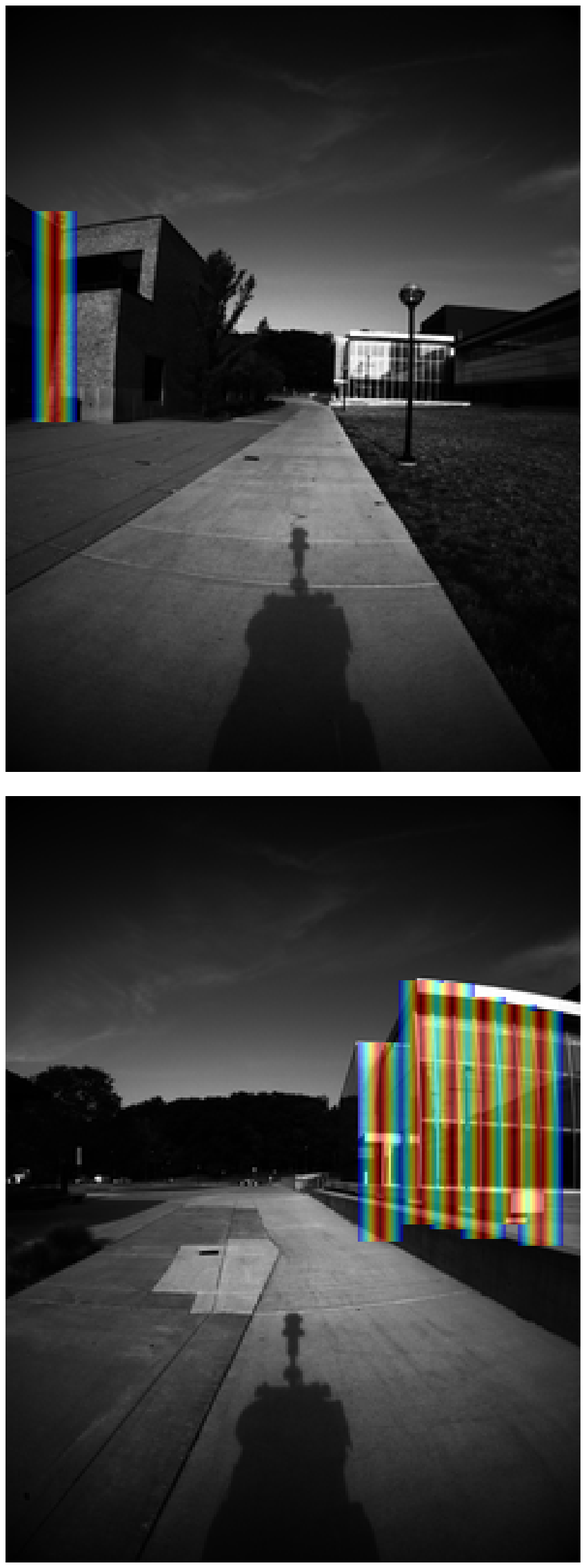}{(a)}\hspace*{-2mm}%
\begin{minipage}[b]{5cm}
\centering
\FIG{2}{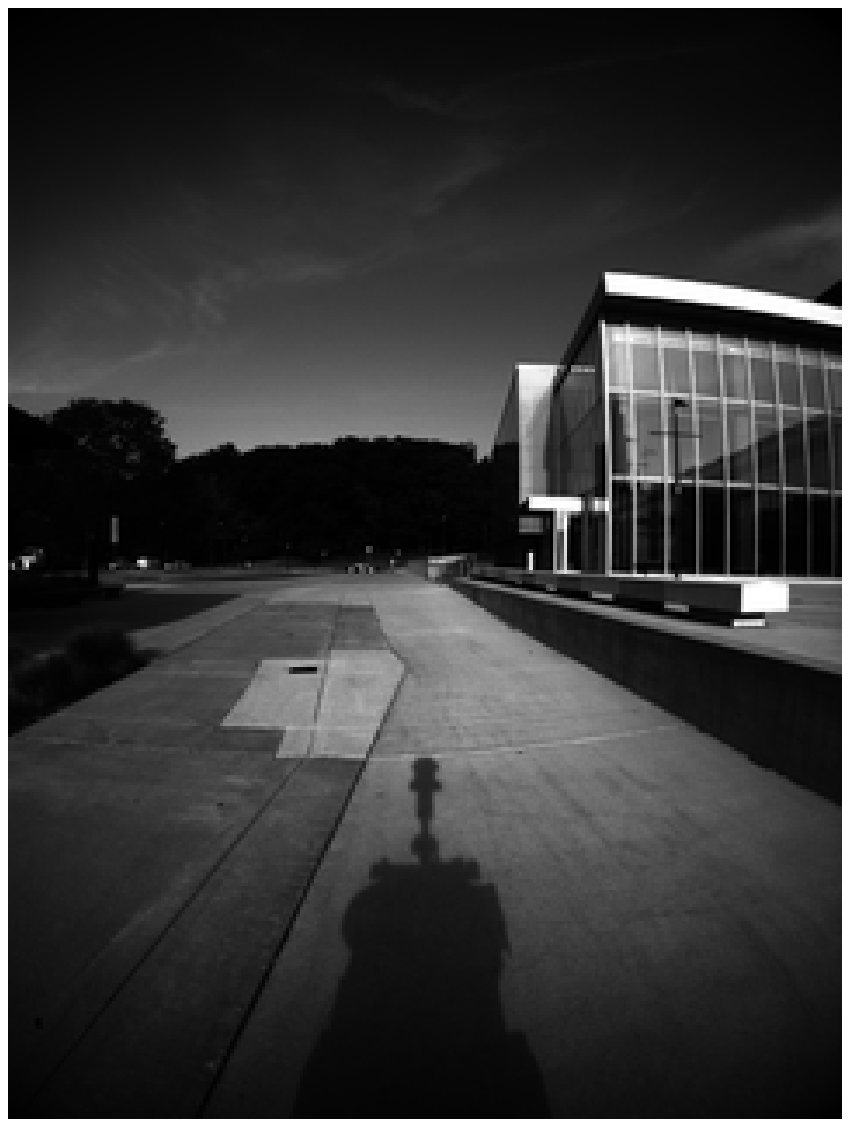}{(b)}
\FIG{2}{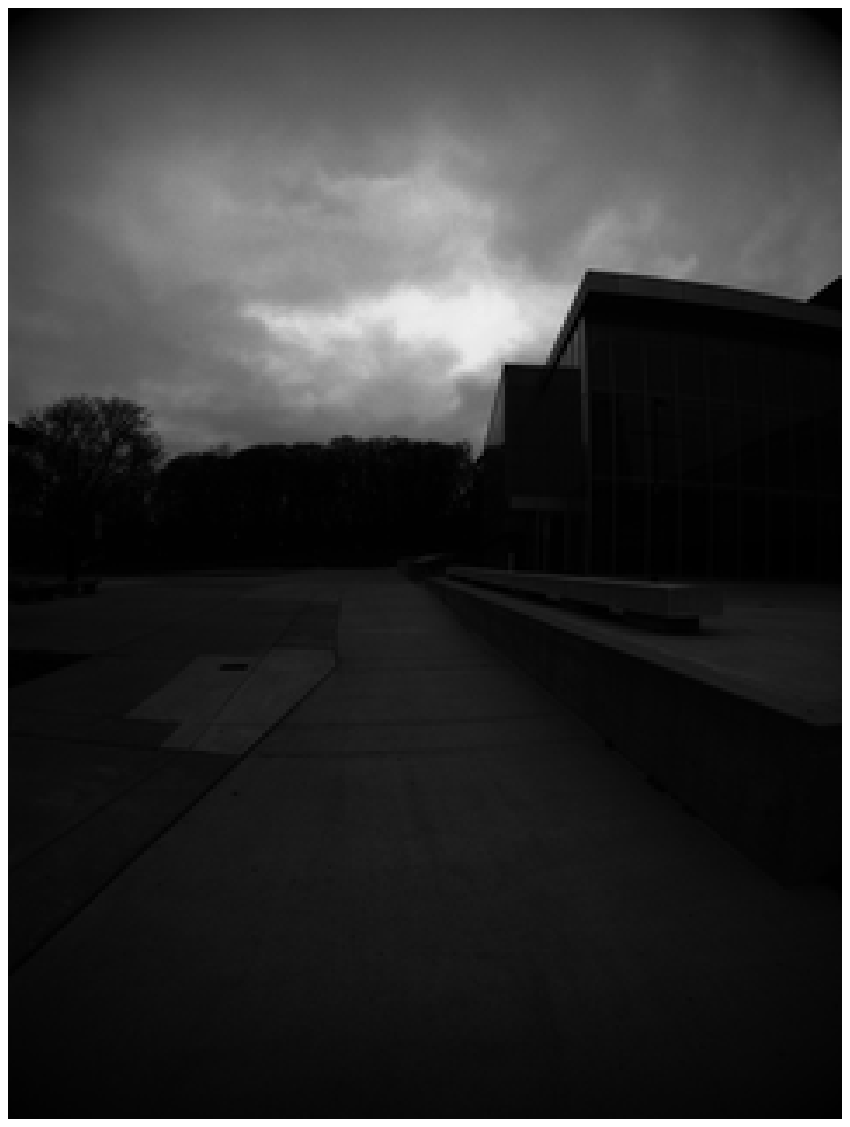}{(c)}\vspace*{-7mm}\\
\FIGx{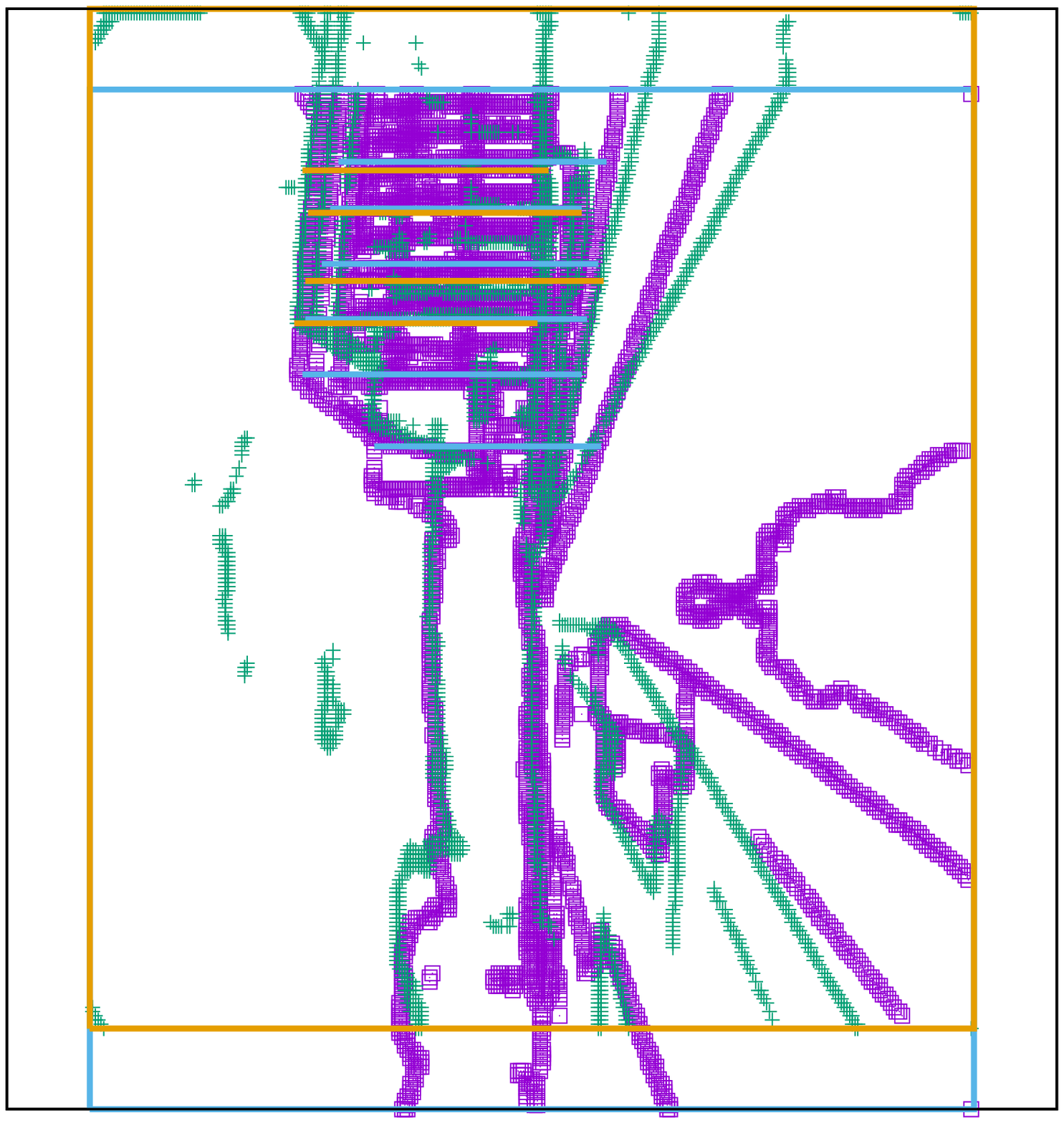}\\
{\scriptsize (d)}
\end{minipage}
\caption{The
domain-invariant next-best-view (NBV) planner
takes a live view image 
and
plans the NBV
that maximizes the possibility of
true-positive detection
of pole-like landmarks,
which are then
used as
the invariant coordinate system
for spatial BoW scene model and matching.
(a)
An input live image (top)
and an NBV view at the planned viewpoint (bottom).
A query view (b) and
a reference view (c)
are aligned
with the invariant coordinate system (d).
}\label{fig:O}
\vspace*{-5mm}
\end{figure}
}

\newcommand{\figG}{
\begin{figure}
\centering
\FIG{8.5}{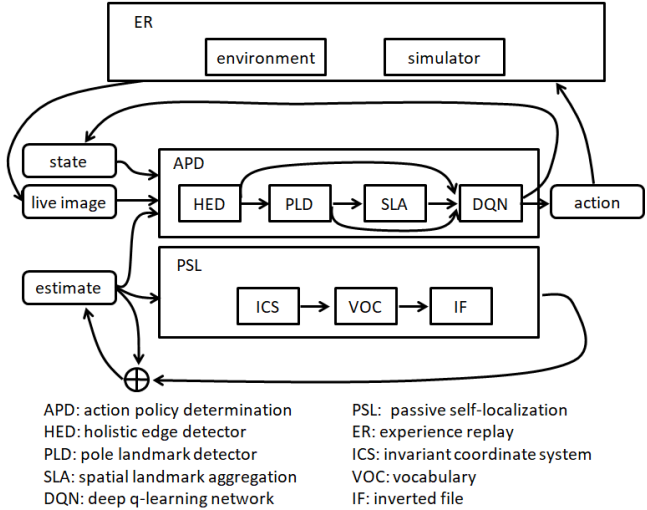}{}
\caption{System architecture.}\label{fig:G}
\end{figure}
}

\newcommand{\EdgeDetector}{
\begin{flushleft}
$
y_1=DoubleConv(x);~~
s_1=SingleConv(x);
$\\$
y_2=DoubleConv(y_1);~~
d_2=MaxPool(y_2);~~
x_2=d_2+s_1;~~
s_2=SingleConv(x_2);
$\\$
p_3=SingleConv(d_2);~~
y_3=DenseBlock(x_2, p_3);~~
d_3=MaxPool(y_3);~~
x_3=d_3+s_2;~~
s_3=SingleConv(x_3);
$\\$
p^2_4=P^2(d_2);~~
p^4=P^4(p^2_4+D_3);~~
y_4=DenseBlock(x_3, p^4);~~
d_4=MaxPool(y_4);~~
x_4=d_4+s_3;~~
s_4=SingleConv(x_4);
$\\$
p^2_5=SingleConv(p^2_4);~~
p_5=SingleConv(b^2_5+d_4);~~
y_5=DenseBlock(x_4, p_5);~~
x_5=y_5+s_4;
$\\$
p_6=SingleConv(y_5);~~
y_6=DenseBlock(x_5, p_6);
$\\$
y_7=Concat(x{\times}1, y_1{\times}1, y_2{\times}2, y_3{\times}4, y_4{\times}8, y_5{\times}16, y_6{\times}16);
$\\$
x_7=SingleConv(y_7);~~
y_7=Aggregate(x_7);
$\\$
x_8=Activate(y_7);~~
y_8=VectorQuantize(y);
$
\end{flushleft}
}

\newcommand{\figE}{
\begin{figure*}
\centering
\FIG{15}{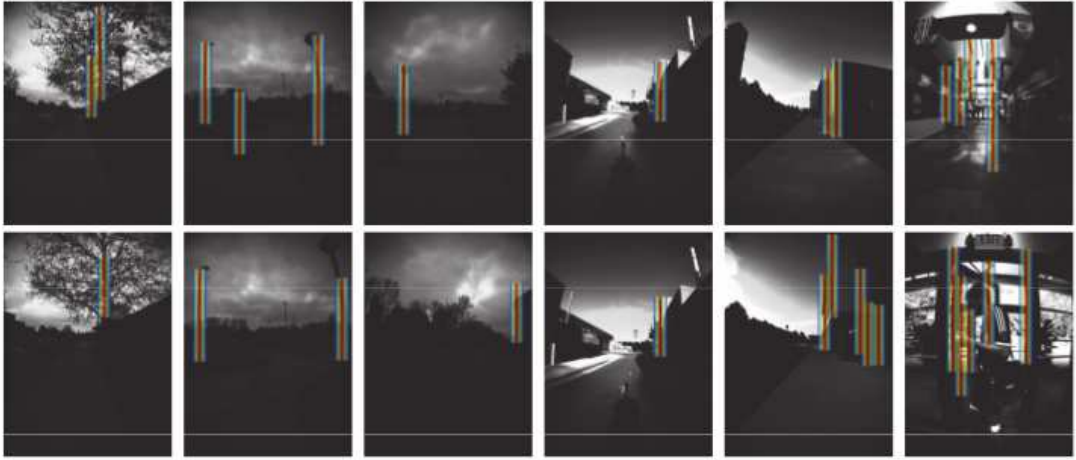}{}
\caption{NBV planning results.
In each figure,
the top and bottom panels
show
the view image
before and after
the planned movements,
respectively.}\label{fig:E}
\end{figure*}
}

\newcommand{\figB}{
\begin{figure}
\FIGR{8}{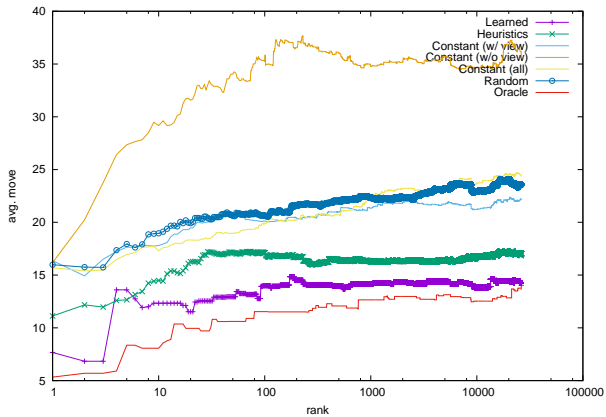}{}
\caption{Cost vs. performance. Vertical axis (cost): travel distance between the initial and final viewpoints. Horizontal axis (performance): ground-truth ranking of the final estimate.}\label{fig:B}
\end{figure}
}

\newcommand{\acprPaperID}{25}

\newcommand{\FIGx}[1]{\includegraphics[height=4.8cm,width=6.4cm,angle=-90]{#1}}

\title{Domain-invariant NBV Planner for Active Cross-domain Self-localization}
\author{Tanaka Kanji 
\thanks{Our work has been supported in part by 
JSPS KAKENHI 
Grant-in-Aid 
for Scientific Research (C) 17K00361, and (C) 20K12008.}
\thanks{K. Tanaka is with Faculty of Engineering, University of Fukui, Japan. 
{\tt\small tnkknj@u-fukui.ac.jp}}
\thanks{We would like to express our sincere gratitude to Koji Takeda and Kanya Kurauchi for development of deep learning architecture, and initial investigation on deep reinforcement learning on the dataset, which helped us to focus on our pole-like landmark project.}}

\maketitle{}

\begin{abstract}
Pole-like landmark has received increasing attention as a domain-invariant visual cue for visual robot self-localization across domains (e.g., seasons, times of day, weathers). However, self-localization using pole-like landmarks can be ill-posed for a passive observer, as many viewpoints may not provide any pole-like landmark view. To alleviate this problem, we consider an active observer and explore a novel ``domain-invariant" next-best-view (NBV) planner that attains consistent performance over different domains (i.e., maintenance-free), without requiring the expensive task of training data collection and retraining. In our approach, a novel multi-encoder deep convolutional neural network enables to detect domain invariant pole-like landmarks, which are then used as the sole input to a model-free deep reinforcement learning -based domain-invariant NBV planner. Further, we develop a practical system for active self-localization using sparse invariant landmarks and dense discriminative landmarks. In experiments, we demonstrate that the proposed method is effective both in efficient landmark detection and in discriminative self-localization.
\end{abstract}

\section{Introduction}

Cross-domain self-localization is the problem of estimating a robot pose given a history of sensor readings including visual images/odometry data, 
with respect to an environment map that was previously collected in different domains (e.g., weathers, seasons, times of the day). A large body of self-localization literature focuses on designing or training a landmark detector that is robust against domain shifts \cite{feat2,Arandjelovic16,feat1}.  
However, the problem of designing/training such a domain-invariant landmark detector is essentially ill-posed when current live images are from a previously unseen domain. Existing solutions can be negatively influenced by environmental and optical effects, such as occlusions, dynamic objects, confusing features, illumination changes, and distortions. One promising approach to address this issue is to utilize inherently invariant landmark objects that are stable and visually invariant against domain-shifts, such as 
walls \cite{wall}, roads \cite{roads}, and poles \cite{poles}. In this study, we are particularly interested in the use of pole-like landmarks because they are ubiquitous both in indoor and outdoor environments.

\figO

Thus far, most of previous works on self-localization suppose a passive observer (i.e., robot), and do not take into account the issue of viewpoint planning, or controlling the observer. However, the pole-like landmark -based self-localization can be ill-posed for a passive observer, as many viewpoints may not provide any pole-like landmark view. Therefore, we consider an active self-localization task that can adapt its viewpoint trajectory, avoiding non-salient scenes that provide no pole-like landmark view, or moving efficiently towards places which are most informative, in the sense of reducing the sensing and computation costs. This is most closely related to the next-best-view (NBV) problem studied in machine vision literature \cite{nbv_org}. However, in our cross-domain setting, a difficulty arises from the fact that the NBV planner is trained and tested in different domains. Existing NBV methods that do not take into account domain shifts would be be confused and deteriorated by the domain-shifts, and require significant efforts for adapting them to a new domain. 

In this work, we propose a novel class of NBV planner, termed ``domain-invariant" NBV planner (Fig. \ref{fig:O}), that attains consistent performance over different domains, without requiring the expensive task of training data collection and retraining (i.e., maintenance-free).
The domain-invariance can be considered as a novel advantage of the proposed approach to existing NBV planners.
Intuitively,
a domain-invariant NBV planner
could restrict itself to take
domain-invariant visual features as 
the sole input.
In addition,
it could
learn domain-invariant statistics of
landmark configuration,
such as
average travel distance between
pole-like landmark views.
Moreover,
it could arrange the plan
depending on
whether
the current live image is
a landmark view or not.
However,
training 
a domain-invariant landmark detector
as well as
an NBV planner,
with reasonable computational cost,
from an available visual experience,
is a non-trivial problem
and which is the focus of our study.

We tackle this problem
by
introducing 
a deep convolutional neural network (DCN) -based landmark detector
and
a deep reinforcement learning (DRL) -based NBV planner.
Our DCN module is built on
a recently developed
multi-scale multi-encoder deep convolutional neural network \cite{hed}. 
Our formulation of DRL is
related
to recent works on deep reinforcement learning -based 
NBV planner \cite{bmvc_nbv}.
However, rather than using typical sensor inputs such as raw images, we propose to use domain-invariant perceptual information,
which is made available by the proposed landmark detector,
to make our NBV planner 
a domain-adaptive one. 

Our main contributions are summarized as follows:
(1) This work is the first to address the issue of domain-invariant NBV using invariant pole-like landmarks. 
(2) Our landmark detection network is built on the recent paradigm in image edge detection,
deep holistic edge detection (HED),
which can detect a small number of holistic essential scene structure, 
rather than 
conventional 
detectors such as Canny \cite{canny}
that tend to detect lots of complex primitive edges.
(3)
The effectiveness of the planning algorithm in terms of self-localization performance and domain-invariance is verified via experiments using publicly available NCLT dataset.

\figA

\section{Problem}

Active self-localization task consists of four distinctive stages: planning, action, detection, and matching (Fig. \ref{fig:A}).
The planning stage
determines the NBV action,
given  
a history of live images and states,
using an NBV planner 
that is trained in a past domain.
The action stage executes the planned NBV action.
The detection stage 
aims 
to detect 
domain-invariant landmark objects
using the pre-trained detector.
The matching stage
aims to query
a pre-built database 
of landmark views
associated with ground-truth GPS viewpoints
to infer the robot pose,
and 
this stage is performed only when one or more landmark objects are detected.
The inference result and all the inference history from previous iterations are integrated to obtain an estimate of the robot pose and its confidence score.
The above four stages are iterated until the confidence score exceeds a pre-set threshold.

\section{Approach}

This section presents the proposed 
framework 
for active visual self-localization (Fig. \ref{fig:G}).
The domain-invariant pole-like landmark detection network (PLD), as well as the domain-adaptive action policy determination block (APD), are emphasized. On the other hand, spatial landmark aggregation module (SLA) can reduce the dimension of PLD output and computational complexity of the DRL -based learner/planner. On the other hand, through fusing invariant pole-like landmarks and discriminative holistic edge images, the passive self-localization block (PSL) can make use of the information from PLD. One the other hand, 
the experience replay (ER) module provides 
random access
to past experiences. 
As a result, these modules or components with each other boost the performance of active self-localization system.

\subsection{NBV Planner}

We use deep Q-learning network (DQN) as the basis for our NBV planner. In our view, the model-free property of Q-learning is desirable for the robotic self-localization applications, as it does not include model parameters to be learned and it can intelligently avoid the possibility of learning a wrong model. However, classical Q-learning fails for large-scale problems due to 
``curse of dimensionality". The recent paradigm of DQN addresses this issue by approximating the value-action function with a deep neural network. Our approach extends this technique in two ways. First, a part of the deep neural network is replaced with a multi-scale multi-encoder DCN block,
derived from the state-of-the-art HED network \cite{hed},
which is very different and significantly more complex than 
existing deep neural network employed
by typical DQN approaches. Second, 
the landmark detector
is pretrained on Big data
before being introduced
as a component of 
the NBV planner.

\figG

\subsection{Pole-like Landmark Detection (PLD)}

Our PLD network architecture is inspired by the recent paradigm  
in the field of holistic edge detection (HED) \cite{hed}. In particular, we are based on the multi-scale multi-encoder HED network that consists of multiple encoder blocks that act as multi-scale encoder blocks. These main blocks consist of 3$\times$3 convolutional sub-blocks that are densely interconnected by the output of the previous main block, and the outputs of each sub-block are connected between them via a 1$\times$1 convolutional block.
The pseudo code of the PLD network $y=f(x)$ is as follows:
\EdgeDetector{}
The operations
'${\times}$' and '$+$' are
the upsampling and addition.
DenseBlock, DoubleConv and SingleConv are
the dense block, double and single convolutional blocks.
MaxPool, 
Concat
are
respectively,
the max pooling,
and 
concatenate operations.
Our implementation basically follows the state-of-the-art HED framework in \cite{hed}.
However,
the HED framework is
modified to deal with our DQN task.
That is,
the 2D edge feature map output by the HED part is spatially aggregated
to produce a horizontal 1D image
that represents the likelihood of
pole-like landmarks ({\it Aggregate}).
Then,
the spatially-aggregated 1D image is
further
input to
the SLA block that consists of 
a pre-trained activation network ({\it Activate})
followed by
a vector quantization 
blocks ({\it VectorQuantize})
to produce an
extremely compact
4-dim landmark feature map.
The aggregation network
summarizes
the likelihood of
pole-like landmark 
in each different horizontal range in the input image, 
respectively
$[0, W/4-1]$,
$[W/4, W/2-1]$,
$[W/2, 3W/4-1]$,
and
$[3W/4, W]$,
where
$W$ is the original image width.
Then,
the 4-dim vector is passed to
the activation network.
The activation network is trained
only once,
in a specific past domain.
This 4-dim vector is input to the value-action function block,
which is described above.
While the locations of pole endpoints in the single train domain are manually annotated in the current study,
unsupervised annotation by introducing techniques like pole-based SLAM \cite{poles}
would be an interesting direction of future research.

\subsection{Implementation Details}

We adopt the spatial bag-of-words (BoW) image model for our PSL module. A BoW model represents every query/mapped image with a collection of vector quantized local visual features, called visual words \cite{video_google}, which are then efficiently indexed/retrieved using an inverted file system. The spatial BoW (SBoW) 
is an extension of the 
BoW image model
to include spatial information of visual words \cite{sbow}.
We observe that the SBoW is applicable
to our case
by
using the horizontal location of pole-like landmark $(x_o, y_o)$
as the domain-invariant reference for spatial information.
More specifically,
we compute
the spatial information of
a given visual feature's location $(x, y)$
as $(x', y')=$ $(x-x_o$, $y)$
(Fig. \ref{fig:O}c). 

A bottleneck of the SBoW is the computation cost to simulate the SPL tasks. That is, the cost for querying the SPL image database for each visual image in each query, is in the order of $O(N_{images}N_{words})$ where $N_{images}$ is the number of visual images per experience and $N_{words}$ is the number of visual words per image. To reduce the cost, it is essential to represent the visual experience in a time/spatial efficient and random accessible manner. To address this, we summarize all the possible PSL tasks in a lookup table, which is then used to simulate the experience replay in an efficient manner. Given a pairing of query and map image sets, the lookup table is constructed in the following procedure. First, the SBoW-based information retrieval system is queried for each query image, and top-$K$ retrieval results are obtained. Then, the similarity between the query image and each of the top-$K$ similar map images is computed. Then, the similarity value is compactly represented in a short $B$-bit distance value. The parameters $K$ and $B$ are empirically set $K=1000$ and $B=8$ considering the accuracy-compactness tradeoff.

The deep Q-learning network is trained in the experience replay procedure.
An experience is defined as
a history of
planned actions,
acquired live images,
and
states.
The NBV action $a$ for the current state $s$ is selected
with the probability in proportional to the function $\exp(Q_{a,s})$.
Our scheme could be replaced with more advanced experience replay frameworks, such as prioritized experience replay \cite{prioritized_experience_replay} 
or dueling network \cite{dueling_q_learning}, 
which would be a future direction of our research.

A navigation task is terminated 
if 
the score distribution
becomes 
a sufficiently narrow distribution.
More formally,
it is judged
if 
the score of the highest scored location
exceeds 
that of the second scored location
with a large margin (0.1).

\figE

\section{Experiements}

We evaluated the effectiveness of the proposed algorithm via 
active cross-domain self-localization in different domains. We used the publicly available NCLT dataset \cite{nclt}. The NCLT dataset is a large-scale, long-term autonomy dataset for robotics research collected at the University of Michigan's North Campus by a Segway vehicle robotic platform. The data we used in the research includes view image sequences along vehicle's trajectories acquired by the front facing camera of the Ladybug3 as well as the ground-truth GPS viewpoint information. It involves both indoor and outdoor change objects such as cars, pedestrians, construction machines, posters, and furniture, during seamless indoor and outdoor navigations of the Segway robot. In our experiments, three different pairings of training and test datasets are used: (2012/1/22, 2012/3/31), (2012/3/31, 2012/8/4), (2012/8/4, 2012/11/17). The four datasets ``2012/1/22", ``2012/3/31", ``2012/8/4", and ``2012/11/17" consist of
26208,
26364,
24138,
and
26923
images.
Images are
resized to 320$\times$240. 

Figures \ref{fig:O} and \ref{fig:E} show examples of pole-like landmark detection.
As shown,
the proposed detector
successfully detected
pole-like landmarks,
such as
poles,
as well as 
vertical edges of building, walls, and doors.
On the other hand,
it can intelligently avoided
false positive detection
of non-structural non-dominant edges,
such as
tiles on the roads
and
leafs on the trees.

Figure \ref{fig:E}
shows
examples of
views before and after
planned NBV actions.
Intuitively
convincing
behaviour
of the robot
was observed.
When a pole-like landmark is near from the robot viewpoint,
the robot tends to plan to move a short distance
so as to avoid lose track of the already observed landmarks.
Otherwise,
the robot tends to plan to move a long distance
so as
to detect unseen landmarks
or approach to already seen landmarks.
Such behaviors are intuitively appropriate
and effective to seek and track landmarks
when one got lost.
It is noteworthy
that 
our approach enables to
learn
such appropriate step sizes
from the available visual experience.

\figB

Figure \ref{fig:B} shows cost vs. accuracy results for quantitative evaluation.
The cost is measured in terms of the number of iterations to reach the termination condition of the active self-localization task, i.e., the confidence score exceeds the pre-set threshold, which is a function of the number of positive detections of pole-like landmark views.
The accuracy is measured in terms of the rank assigned to the ground-truth robot pose when the active self-localization task is terminated.

Our proposed method (``Learned") is compared against six different methods: 
``Heuristics",
``Constant (w/ view)",
``Constant (w/o view)",
``Constant (all)",
and
``Oracle".
``Heuristics"
is a manually designed strategy
that
selects either of 
short or long distance move,
based on a pre-designed heuristics.
Intuitively,
long distance move
is useful to search
for out-of-view landmark objects,
while
short distance move
is useful to approach 
to
landmark objects in the field of view
to obtain 
a closer view of them.
Based on the idea,
the heuristics
selects long distance move
$C_{long}+\Delta d$
if there is no detected landmark in the live image,
or
short distance move
$C_{short}+\Delta d$
otherwise,
where
$\Delta d$ is an artificial motion noise in range $[-1, 1]$ [m].
$C_{long}$
and
$C_{short}$
are two different constant
parameters 
that are learned in the training domain.
To learn
the constants $C_{short}$ and $C_{long}$,
the training image dataset is
split into two disjoint subsets,
i.e.,
images with and without pole-like landmark views,
and
then
$C_{short}$ and $C_{long}$,
is optimized to
maximize 
for individual subsets
the possibility of positive landmark detection.
The next three methods,
``Constant (w/ view)",
``Constant (w/o view)",
and
``Constant (all)"
determine the robot step size,
respectively,
as
$C_{short}\pm 1$,
$C_{long}\pm 1$,
and
$C_{all}\pm 1$,
regardless of
whether the live image is pole-like landmark view or not.
$C_{all}$
is learned in the training domain,
in a similar procedure as in
$C_{short}$
and
$C_{long}$,
but by using 
all the training images
instead of the subsets.
Cost versus performance for the four pairings of training and test seasons are
shown in Fig. \ref{fig:B}.
It is clear
that 
the proposed approach 
significantly outperforms
all the approaches considered here.
It could be concluded that 
a non-trivial good policy 
was learned with a model-free DQN
from
only domain-invariant landmarks and HED images.

\section{Conclusions}

In this paper, a novel framework of cross-domain active self-localization using pole-like landmarks 
was proposed.
Unlike previous approaches to active self-localization,
we hypothesize that
pole-like landmarks are inherently invariant, stable and ubiquitous cue for visual robot self-localization.
Based on the idea,
a novel multi-scale multi-encoder landmark detection network is introduced to enable detection of holistic essential landmarks,
which 
are then used as the sole input to an off-policy  model-free 
DQN-based
NBV planner.
The result is a domain-invariant variant of NBV planner that attains consistent performance over different domains, without requiring the expensive task of training data collection and retraining. The effectiveness of the planning algorithm in terms of self-localization performance and domain-invariance 
was 
experimentally verified using publicly available NCLT dataset.

\bibliographystyle{IEEEtran}
\bibliography{rl20}

\end{document}